\documentclass[final,1p,times]{elsarticle}

\usepackage{lipsum}
\makeatletter
\def\ps@pprintTitle{%
 \let\@oddhead\@empty
 \let\@evenhead\@empty
 \def\@oddfoot{}%
 \let\@evenfoot\@oddfoot}
\makeatother

\usepackage{graphicx}
\usepackage{amssymb}
\usepackage{amsmath}

\usepackage{float}

\begin{document}

\begin{frontmatter}

\title{Unsupervised Contextual Anomaly Detection using Joint Deep Variational Generative Models}
\author{Yaniv Shulman}
\address{yaniv@aleph-zero.info}

\begin{abstract}
A method for unsupervised contextual anomaly detection is proposed using a cross-linked pair of Variational Auto-Encoders (VAE) for assigning a normality score to an observation. The method enables a distinct separation of contextual from behavioral attributes and is robust to the presence of anomalous or novel contextual attributes. The method can be trained with data sets that contain anomalies without any special pre-processing.
\end{abstract}
\end{frontmatter}

\section{Introduction}
\label{S:Introduction}

Anomaly detection is an important area of research since anomalies represent a substantial deviation from the normal characteristics of a system or process of interest. Often these processes result in highly dimensional data sets, with complex relationships within the data and exhibit stochastic behavior. Furthermore the anomalies by definition contain high self-information measure and therefore carry useful information about the underlying data generation process. There exist a number of similar definitions of what an anomaly is however in this paper the following definition is adopted \cite{CompareAnomalyDetection2016}:
\begin{enumerate}
\item Anomalies are different from the norm in respect to their attributes.
\item They are rare in a data set compared to the normal instances.
\item In addition a \textit{novel} observation is defined as an observation that is substantially different than any observation in the training data set.
\end{enumerate}

In this paper a method for contextual anomaly detection is proposed using a cross-linked pair of Variational Auto-Encoders (VAE) for assigning a normality score to an observation. The method enables a distinct separation of contextual from behavioral attributes and is robust to the presence of anomalous or novel contextual attributes. The method can be trained with data sets that contain anomalies without any special pre-processing. In addition the method can be extended in a straight forward way to further decompose and separately model the joint variational approximation by introducing additional independent recognition networks thus allowing for more accurate representation in the latent space.
\newpage

In summary the key contributions of this paper are:
\begin{itemize}
\item A novel architecture for auto-encoding joint latent variational Bayes.
\item A novel method for robust unsupervised anomaly detection in the presence of contextual anomalies.
\end{itemize}

\section{Preliminaries}
\label{S:Preliminaries}

\subsection{Anomaly Detection}
\label{S:Preliminaries:Anomaly Detection}
In this section a number of criteria for broadly categorizing anomaly detection algorithms is briefly discussed. These concepts are covered in more detail in \cite{Aggarwal:2013:OA:2436823, Chandola:2009, CompareAnomalyDetection2016}.
\newline
\newline
\textit{Proximity based anomaly detection} assumes that anomalous data are isolated from the majority of the data whether in relation to clusters or global/local dense regions. To determine if an observation is anomalous, the distance to the clusters or the density estimate is calculated to generate a normality score; \textit{Statistical  based anomaly detection} assumes that data is generated from a known probability distribution which can be described by parametric or non-parametric formulation. To determine if a data point is an anomaly the probability of it being generated from the assumed distribution is determined and a normality score is produced derived from this probability; \textit{Deviation based anomaly detection} is based on the reconstruction errors following a spectral or other transformation of the data to a lower dimensional space and then back to the original space. The magnitude of the reconstruction error is used to generate a normality score.
\newline
\newline
\textit{Supervised anomaly detection} is employed where both the training and test data sets specify for each observation whether it is normal or anomalous; \textit{Semi-supervised anomaly detection} is typically defined as scenarios where the training data contains only normal observations; \textit{Unsupervised anomaly detection} is the case where there are no labels provided in either the training or the testing data sets and no assumptions are made on the existence or number of anomalous observations in the available data.
\newline
\newline
\textit{Contextual anomaly detection} is formulated such that the data contains two types of attributes, behavioral and contextual attributes. \textit{Behavioral attributes} are attributes that relate directly to the process of interest whereas \textit{contextual attributes} relate to exogenous but highly affecting factors in relation to the process. Generally the behavioral attributes are conditional on the contextual attributes.

\subsection{Variational Auto-Encoder}
\label{S:Preliminaries:Variational Auto-Encoder}
In this section a brief overview of the Variational Auto-Encoder (VAE) \cite{Kingma:2009} is provided for presenting the notation used in subsequent sections of the paper.
\newline
A Variational Auto-Encoder (VAE) is a directed probabilistic graphical model that enables an efficient variational inference for intractable posterior distributions which are approximated by a neural network. The VAE is comprised of two serially adjoined neural networks which are referred to as encoder/recognizer and decoder/generator respectively. The generator network $g(\mathbf{z}, \theta)$ where $\mathbf{z}$ is a latent variable approximates the generative process $p_{\theta}(\mathbf{x})=p_{\theta}(\mathbf{x}|\mathbf{z})p_{\theta}(\mathbf{z})$. The recognition network $f(\mathbf{x}, \phi)$ models $q_{\phi}(\mathbf{z}| \mathbf{x})$ a variational approximation of the intractable posterior $p_{\theta}(\mathbf{z}| \mathbf{x})$. All parameters are learned jointly and efficiently by employing the Stochastic Gradient Variational Bayes (SGVB) \cite{Kingma:2009} estimator. As the marginal likelihood of the data $p(\textbf{x})$ is intractable, the problem is transformed into an optimization problem where the objective function of the VAE is the Evidence Lower Bound (ELBO), a lower bound on $log\,p(\textbf{x})$ as formulated in equation \ref{eq:vae:elbo_c}.

\begin{subequations}
\begin{align}
log\,p_{\theta}  ( \{x^{(i)}\}_{i=1}^{N}) &= \sum_{l=1}^{N} log\,p_{\theta}(x^{(i)}) \label{eq:vae:marginal_likelihood_of_evidence_a} \\
&= KL(q_{\phi}(\mathbf{z}|\mathbf{x}) \parallel p_{\theta}(\mathbf{z})) + \mathcal{L}(\theta, \phi, \mathbf{x}) \label{eq:vae:marginal_likelihood_of_evidence_b}
\end{align}
\end{subequations}

\begin{subequations}
\begin{align}
log\,p_{\theta}(\mathbf{x}) &\geq \mathcal{L}(\theta, \phi, \mathbf{x}) \label{eq:vae:elbo_a} \\
&= \mathbb{E}_{q_{\phi}(\mathbf{z}|\mathbf{x})}[-log\,q_{\phi}(\mathbf{z}|\mathbf{x}) + log\,p_{\theta}(\mathbf{x} | \mathbf{z})] \label{eq:vae:elbo_b} \\
&= -KL(q_{\phi}(\mathbf{z}|\mathbf{x}) \parallel p_{\theta}(\mathbf{z})) + \mathbb{E}_{q_{\phi}(\mathbf{z}|\mathbf{x})}[log\,p_{\theta}(\mathbf{x}|\mathbf{z})] \label{eq:vae:elbo_c}
\end{align}
\end{subequations}

Where the inequality in equation \ref{eq:vae:elbo_a} follows from the non-negativity of the Kullback–Leibler divergence. A complete derivation can be found in \cite{doi:10.1080/01621459.2017.1285773}.

\subsection{Conditional Variational Auto-Encoder}
\label{S:Preliminaries:Conditional Variational Auto-Encoder}
In this section a very brief overview of the Conditional Variational Auto-Encoder (CVAE) \cite{NIPS2015_5775}. The CVAE expands on the learning capacity of the VAE by defining an architecture that enables the model to learn explicit joint variational approximation of the latent variable $q_{\phi}(\mathbf{z} | \mathbf{x}, \mathbf{y})$ and a directly modulated conditional generative $p_{\theta}(\mathbf{y} | \mathbf{x}, \mathbf{z})$ model. In CVAE the input is denoted as $\mathbf{x}$, the output is denoted as $\mathbf{y}$ and the latent variable is $\mathbf{z}$. The CVAE utilizes the SGVB optimization framework and an objective function closely related to the VAE defined in equation \ref{eq:cvae:elbo_c}.

\begin{subequations}
\begin{align}
&log\,p_{\theta}(\mathbf{y}|\mathbf{x}) = \\ 
&KL(q_{\phi}(\mathbf{z}|\mathbf{x},\mathbf{y}) \parallel p_{\theta}(\mathbf{z}|\mathbf{x}, \mathbf{y})) + \mathbb{E}_{q_{\phi}(\mathbf{z}|\mathbf{x}, \mathbf{y})}[-log\,q_{\phi}(\mathbf{z}|\mathbf{x}, \mathbf{y}) + log\,p_{\theta}(\mathbf{y},\mathbf{z}|\mathbf{x})] \\
&\geq \mathbb{E}_{q_{\phi}(\mathbf{z}|\mathbf{x}, \mathbf{y})}[-log\,q_{\phi}(\mathbf{z}|\mathbf{x}, \mathbf{y}) + log\,p_{\theta}(\mathbf{y},\mathbf{z}|\mathbf{x})] \\
&=\mathbb{E}_{q_{\phi}(\mathbf{z}|\mathbf{x}, \mathbf{y})}[-log\,q_{\phi}(\mathbf{z}|\mathbf{x}, \mathbf{y}) + log\,p_{\theta}(\mathbf{z}|\mathbf{x})] + \mathbb{E}_{q_{\phi}(\mathbf{z}|\mathbf{x}, \mathbf{y})}[log\,p_{\theta}(\mathbf{y}|\mathbf{x}, \mathbf{z})] \\
&=-KL(q_{\phi}(\mathbf{z}|\mathbf{x},\mathbf{y}) \parallel p_{\theta}(\mathbf{z}|\mathbf{x})) + \mathbb{E}_{q_{\phi}(\mathbf{z}|\mathbf{x}, \mathbf{y})}[log\,p_{\theta}(\mathbf{y}|\mathbf{x}, \mathbf{z})] \label{eq:cvae:elbo_c}
\end{align}
\end{subequations}

\subsection{Related Work}
\label{S:Preliminaries:Conditional Variational Auto-Encoder}
Anomaly detection  has attracted large interest from the research community over decades due to the varied areas of application and theoretical importance. There are many suggested methods for the general case however a much smaller number of methods that deal explicitly with contextual anomaly detection exist. A review of related work is given in \cite{Chandola:2009} and in \cite{CompareAnomalyDetection2016}, the latter being more recent and also endeavors to provide an elaborate comparative evaluation for a large number of methods. In this section the focus is on more recent methods proposed either for contextual anomaly detection or anomaly detection that make use of variational inference and deep learning methods. Note that both supervised, semi-supervised and unsupervised methods are included. \cite{DBLP:journals/corr/LiangP16} proposes a contextual anomaly detection method (ROCOD) for dealing with situations where there are abnormal or sparse contextual attributes by utilizing local and global behavioral models conditional on the context. \cite{DBLP:journals/corr/LiangP16} also performs comparative analysis of a number of methods and demonstrates that state-of-the-art point methods achieve relatively poor results on contextual anomaly detection problems. \cite{Song2007ConditionalAD} has proposed a method for general contextual anomaly detection and proposes three different expectation-maximization algorithms for learning the model. Additionally \cite{Song2007ConditionalAD} comparatively evaluates more than 13 different data sets against several other non-contextual anomaly detection methods. \cite{DBLP:conf/aaai/HongH16} propose a multivariate conditional outlier detection framework for clinical applications by defining a multi-variate function to calculate the normality score. \cite{DBLP:journals/corr/abs-1802-00187} propose a method for improved unsupervised learning of $L^{2}$ constrained representations for clustering analysis using deep Auto-Encoders. Normality scores are then calculated based on similarity measure to clusters. Note that in \cite{DBLP:journals/corr/abs-1802-00187} the number of clusters is assumed to be known. \cite{Slch2016VariationalIF} apply a Stochastic Recurrent Network (STORN) \cite{Bayer:2015} for supervised detection of anomalies in robot sensors time series data. \cite{An2015VariationalAB} suggests an anomaly detection method using a VAE and proposes the \textit{Reconstruction Probability} a novel normality score based on the probabilistic measure expressed in the objective function of the VAE. \cite{Xu:2018:UAD:3178876.3185996} suggest Donot, an unsupervised anomaly detection algorithm utilizing a Variational Auto-Encoder for anomaly detection in Seasonal KPI arising from web applications utilizing the Reconstruction Probability. 

\section{Problem Description}
\label{S:Problem Description}
\subsection{Unsupervised Contextual Anomaly Detection}
\label{S:Preliminaries:Unsupervised Contextual Anomaly Detection}
Most anomaly detection methods known to the author at this time do not provide explicit treatment of contextual and behavioral attributes separately but simply merge the two attribute types into a single observation thus transforming the original task into a standard point anomaly detection \cite{CompareAnomalyDetection2016, Chandola:2009}. On the other hand some contextual anomaly detection methods either require a labeled data set for training or are designed for specific domains therefore it seems not many methods exist to perform general unsupervised contextual anomaly detection. Furthermore by definition relatively little information is available on the distribution of the behavioral attributes in low density areas of the contextual subspace which results in an additional challenge for the existing algorithms especially when there is no information available on the distribution of the behavioral attributes when the context is in itself novel. 
\newline
\newline
In this paper the focus is on unsupervised contextual anomaly detection where the training and testing data sets are generated by the same process.  It is of interest to develop a robust model that is able to learn efficiently the state of the process and correctly predict an observation as an anomaly when the behavioral attributes are in fact an anomaly given the context. However it is desirable for such a model to be robust to anomalies present in the contextual attributes and use the “best” available relevant context to make meaningful predictions.

\newpage

\section{Proposed Method}
\label{S:Proposed Method}
\subsection{The Data Generation Model}
\label{S:Proposed Method:The Data Generation Model}
Given a data set of observations $\mathcal{D} = \{d^{(i)}=[{c}^{(i)}, {x}^{(i)}] \, |\, c^{(i)} \in \mathbf{C} , x^{(i)} \in \mathbf{X} \}_{i=1}^{N}$ where $[\circ,\circ]$ denotes concatenation, the set $\mathbf{X} = \{x^{(i)}\}_{i=1}^{N}$ contains only behavioral attributes and the set $\mathbf{C} = \{c^{(i)}\}_{i=1}^{N}$ contains only the corresponding contextual attributes and $[{x}^{(i)}, {c}^{(i)}]$ are jointly and independently drawn. The data generation process where the $N$ samples are taken can be modeled as follows: 
\begin{enumerate}
\item A sample $z_{x}^{(i)}$ is taken from a latent variable $\mathbf{z_{x}}$ with prior distribution $p_{\theta}(\mathbf{z_{x}})$.
\item A sample $z_{c}^{(i)}$ is taken from a latent variable $\mathbf{z_{c}}$ with prior distribution $p_{\theta}(\mathbf{z_{c}})$.
\item A sample $c^{(i)}$ is taken from a variable $\mathbf{c}$ with conditional distribution $p_{\theta}(\mathbf{c} | \mathbf{z_{c}})$.
\item A sample $x^{(i)}$ is taken from a variable $\mathbf{x}$ with conditional distribution $p_{\theta}(\mathbf{x}|\mathbf{z_{x}}, \mathbf{z_{c}})$.
\end{enumerate}

\begin{figure}[t]
\label{F:Proposed Method:1}
\centering\includegraphics[width=0.5\linewidth]{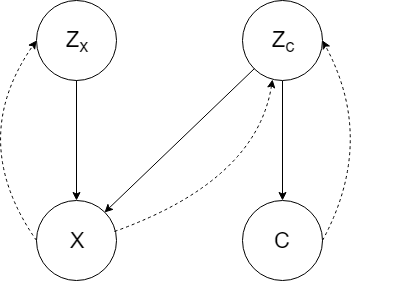}
\caption{Illustration of the generative model as a directed graphical model. $\mathbf{x}$ is the behavioral attributes for the process of interest, $\mathbf{c}$ is the contextual attributes which in this case do not participate directly in the generative process $p_{\theta}(\mathbf{x}) = \iint{p_{\theta}(\mathbf{x}|\mathbf{z_{x}},\mathbf{z_{c}})p_{\theta}(\mathbf{z_{x}})p_{\theta}(\mathbf{z_{c}})d\mathbf{z_{x}}d
\mathbf{z_{c}}}$. Solid lines denote the generative process whereas dashed line denote the variational approximations.}
\end{figure}

The generative process is defined as $p_{\theta}(\mathbf{x}) = \iint{p_{\theta}(\mathbf{x}|\mathbf{z_{x}},\mathbf{z_{c}})p_{\theta}(\mathbf{z_{x}})p_{\theta}(\mathbf{z_{c}})d\mathbf{z_{x}}d\mathbf{z_{c}}}$, 
\newline 
$p_{\theta}(\mathbf{c}) = \int{p_{\theta}(\mathbf{c}|\mathbf{z_{c}})p_{\theta}(\mathbf{z_{c}})d
\mathbf{z_{c}}}$ and is chosen so to prevent $\mathbf{c}$ from modulating the generative process of $\mathbf{x}$ directly for reasons brought in subsequent sections. Figure \ref{F:Proposed Method:1} provides an overview of the generative process.  $p_{\theta}(\mathbf{x})$ and $p_{\theta}(\mathbf{c})$ are often intractable.

\subsection{Joint Deep Variational Generative Models}
\label{S:Proposed Method: Joint Deep Variational Generative Models}
\subsubsection{The Variational Bound}
\label{S:Proposed Method:Learning Objective}
Let $\mathbf{z}=[\mathbf{z_{x}}, \mathbf{z_{c}}]$ denote the complete set of latent variables. The variational lower bound of $p_{\theta}(\mathbf{x})$ and $p_{\theta}(\mathbf{c})$ is defined as follows:

\begin{equation}
\label{eq:jlvae:elbo_c}
log\,p_{\theta}(\mathbf{c}) \geq -KL(q_{\phi}(\mathbf{z_{c}}|\mathbf{x},\mathbf{c}) \parallel p_{\theta}(\mathbf{z_{c}})) + \mathbb{E}_{q_{\phi}(\mathbf{z_{c}}|\mathbf{x},\mathbf{c})}[log\,p_{\theta}(\mathbf{c}|\mathbf{z_{c}})]
\end{equation}

\begin{equation}
\label{eq:jlvae:elbo_x}
log\,p_{\theta}(\mathbf{x}) \geq -KL(q_{\phi}(\mathbf{z}|\mathbf{x}) \parallel p_{\theta}(\mathbf{z})) + \mathbb{E}_{q_{\phi}(\mathbf{z}|\mathbf{x})}[log\,p_{\theta}(\mathbf{x}|\mathbf{z})]
\end{equation}

 To optimize jointly the variational lower bound objective of the two marginal likelihoods equations \ref{eq:jlvae:elbo_c} and \ref{eq:jlvae:elbo_x} are combined.

\begin{equation}
\label{eq:jlvae:elbo_total}
\begin{split}
&log\,p_{\theta}(\mathbf{c}) + log\,p_{\theta}(\mathbf{x}) \geq \\
&-KL(q_{\phi}(\mathbf{z_{c}}|\mathbf{x},\mathbf{c}) \parallel p_{\theta}(\mathbf{z_{c}})) + \mathbb{E}_{q_{\phi}(\mathbf{z_{c}}|\mathbf{x},\mathbf{c})}[log\,p_{\theta}(\mathbf{c}|\mathbf{z_{c}})] \\ &-KL(q_{\phi}(\mathbf{z}|\mathbf{x}) \parallel p_{\theta}(\mathbf{z})) + \mathbb{E}_{q_{\phi}(\mathbf{z}|\mathbf{x})}[log\,p_{\theta}(\mathbf{x}|\mathbf{z})]
\end{split}
\end{equation}

Given the KL terms in equation \ref{eq:jlvae:elbo_total} may be integrated analytically under certain conditions for calculating the empirical loss, the objective is optimized using the Stochastic Gradient Variational Bayes (SGVB) \cite{Kingma:2009} estimator:

\begin{equation}
\begin{split}
\label{eq:jlvae:elbo_total_empirical}
&\mathcal{L}(\theta, \phi, c^{(i)}, x^{(i)}) = \\
&-KL(q_{\phi}(z_{c}^{(i)}|x^{(i)},c^{(i)})\parallel p_{\theta}(z_{c}^{(i)})) -KL(q_{\phi}(z^{(i)}|x^{(i)})\parallel p_{\theta}(z^{(i)})) \\
&+ \frac{1}{L}\sum_{l=1}^{L} log\,p_{\theta}(c^{(i)}|z_{c}^{(i,l)}) + \frac{1}{L}\sum_{l=1}^{L} log\,p_{\theta}(x^{(i)} | z^{(i,l)})
\end{split}
\end{equation}

Where $z_{c}^{(i,l)} = g_{\phi}(x^{(i)}, c^{(i)},\varepsilon_{c}^{(i,l)}), \varepsilon_{c} \sim \mathcal{N}(\mathbf{0}, \mathbf{I})$ and $z^{(i,l)} = h_{\phi}(x^{(i)},c^{(i)},\varepsilon^{(i,l)}), \varepsilon \sim \mathcal{N}(\mathbf{0}, \mathbf{I})$, $L$ is the number of samples. The first two $KL$ terms in equation \ref{eq:jlvae:elbo_total_empirical} represent the latent error for the two variational distributions $q_{\phi}(\mathbf{z_{c}} | \mathbf{x},\mathbf{c})$ and $q_{\phi}(\mathbf{z}| \mathbf{x})$, and the two remaining terms the log probability of the reconstruction errors for the contextual and behavioral attributes $\mathbf{C} = \{c^{(i)}\}_{i=1}^{N}$ and $\mathbf{X} = \{x^{(i)}\}_{i=1}^{N}$ respectively.

\subsubsection{Architecture}
\label{S:Proposed Method:Architecture}
To approximate the posteriors of the joint generative models $p_{\theta}(\mathbf{c}|\mathbf{z_{c}})$ and $p_{\theta}(\mathbf{x}|\mathbf{z_{x}},\mathbf{z_{c}})$ two recognition networks and two generator networks are jointly trained. The behavioral attributes $\mathbf{x}$ are input into one of the recognition networks and both the contextual and behavioral attributes $[\mathbf{x},\mathbf{c}]$ are input into the other. Both recognition networks output the parameters of the variational approximations to the prior followed by $L$ samples that are drawn from the variational approximations to form a Monte Carlo approximation of the expectations of the reconstruction with respect to variational approximations \cite{Kingma:2009}. This architecture provides a number of benefits:

\begin{enumerate}
\item Explicit treatment of behavioral and contextual attributes.
\item Enables an indirect modulation of the generative process of $p_{\theta}(\mathbf{x}|\mathbf{z})$ by $\mathbf{c}$ based on the latent representation of the contextual attributes rather than a direct modulation of the process as done in a CVAE architecture which results in increased robustness to the presence of outliers and novelties in the contextual space. Intuitively this can be explained by the similarity of the latent representation of $\mathbf{c}$ to spectral dimensionality reduction representation which maps the data into a known sub-space, but with the increased model capacity of the recognition network and the benefit of a probabilistic interpretation.
\item Enables assigning different priors for the contextual and behavioral spaces, having multiple of each as a method to decompose and separately model the joint latent distribution.
\end{enumerate}

\subsubsection{Training and Classification}
\label{S:Proposed Method:Training and Classification}
\begin{figure}[t]
\label{F:Proposed Method:2}
\centering\includegraphics[width=0.8\linewidth]{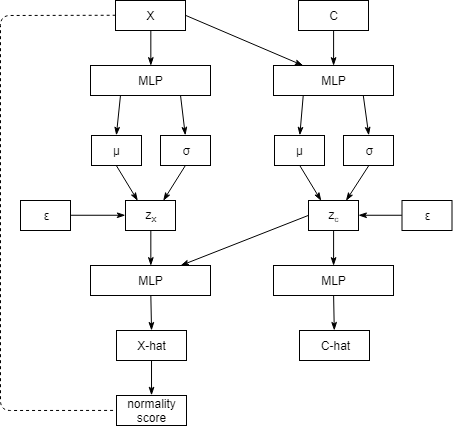}
\caption{Illustration of the architecture}
\end{figure}

All recognition and generator networks are jointly trained using the Stochastic Gradient Variational Bayes (SGVB) \cite{Kingma:2009} estimator. Having learned the model parameters a normality score can be obtained by either calculating a reconstruction error norm for the behavioral attributes $|| x^{(i)} - \hat{x}^{(i)}||$ or by calculating the Reconstruction Probability of  $x^{(i)}$ defined as $\mathbb{E}_{qx_{\phi}(\mathbf{z}|\mathbf{x})} log\,p_{\theta}(\mathbf{x} | \mathbf{z})$ \cite{An2015VariationalAB}. Note that the reconstructed context $\hat{c}^{(i)}$ is not strictly required for assigning a normality score for classification but can be used to estimate the normality score of the context if desired.  Figure \ref{F:Proposed Method:2} provides an overview of the architecture.

\section{Experimental Results}
\label{S:Experimental Results}
\subsection{Kddcup99}
\label{S:Kddcup99}
\subsubsection{Data}
\label{S:Experimental Results:Kddcup99:Data}
Comparative evaluation of contextual anomaly detection methods is a challenging task due to lack of availability of common and suitable data sets that are both labeled and partitioned into behavioral and contextual attributes. To overcome this challenge a publicly available data set the Kddcup99\footnote{http://kdd.ics.uci.edu/databases/kddcup99/kddcup99.html} was adopted as well as an evaluation method used in \cite{DBLP:journals/corr/LiangP16} to provide a performance baseline. The Kddcup99  is "the data set used for The Third International Knowledge Discovery and Data Mining Tools Competition, which was held in conjunction with KDD-99 The Fifth International Conference on Knowledge Discovery and Data Mining. The competition task was to build a network intrusion detector, a predictive model capable of distinguishing between ``bad'' connections, called intrusions or attacks, and ``good'' normal connections. This database contains a standard set of data to be audited, which includes a wide variety of intrusions simulated in a military network environment". The Kddcup99 data set is by a large margin the most challenging data set evaluated by \cite{DBLP:journals/corr/LiangP16} and therefore was elected for this experiment. An effort was made to adhere to the same method of pre-processing and data inclusion as described in \cite{DBLP:journals/corr/LiangP16} however there are some differences as described subsequently. 

The observations from the \textit{r2l} and \textit{u2r} attack families were retained as well as attacks of type \textit{ipsweep} and \textit{nmap}, and normal observations. This results in a total of 605,803 observations out of which 595,797 are labeled as normal, and the rest 10,006 are considered anomalies (approx. 1.652\%). Similarly to \cite{DBLP:journals/corr/LiangP16} the \textit{service}, \textit{duration}, \textit{src\_bytes} and \textit{dst\_bytes} were used as behavioral attributes and all other as contextual attributes. The logarithm of \textit{duration}, \textit{src\_bytes} and \textit{dst\_bytes} was taken since these attributes are processed in the same manner in \cite{DBLP:journals/corr/LiangP16}. All categorical features were one-hot-encoded and finally all attributes are normalized to $[0,1]$ range. The resulting data set contains 65 behavioral attributes and 45 contextual attributes and enables quantitative analysis of the proposed algorithm's effectiveness against the algorithms evaluated in \cite{DBLP:journals/corr/LiangP16} on a similar data set.

\subsubsection{Model}
\label{S:Experimental Results:Kddcup99:Model}
The model is comprised of behavioral recognizer and generator networks and contextual recognizer and generator networks as in the basic architecture described in section \ref{S:Proposed Method:Architecture} and illustrated in figure \ref{F:Proposed Method:2}. The arrangement of units in the behavioral recognizer MLP were: 65 (input), 58, 32 and 4 units for the latent output, with the generator having a mirror architecture. The arrangement of units in the contextual recognizer MLP were: 110 (input), 40, 22 and 4 units for the latent output, with the generator having a mirror architecture except for the output layer containing 45 units. All activation functions in the MLPs are Relu where applicable, however the latent parameters layer as well as the outputs of both generators employ linear activation. Isotropic normal distribution were assumed to the data and latent distributions which lead to the total empirical objective is presented in equation \ref{eq:empirical_loss}, note there is an added L1 regularization term over the MLPs' weights with $\lambda=10^{-5}$.
\begin{equation}
\label{eq:empirical_loss}
\begin{split}
&\mathcal{L}(\theta, \phi, c^{(i)}, x^{(i)}) = \\
&- \frac{1}{2} \bigg[\sum_{i=1}^{|z|}   (1 + log((\sigma_{z}^{(i)})^{2}) - (\mu_{z}^{(i)})^{2} - (\sigma_{z}^{(i)})^{2}) + \sum_{i=1}^{|z_{c}|}   (1 + log((\sigma_{z_{c}}^{(i)})^{2}) - (\mu_{z_{c}}^{(i)})^{2} - (\sigma_{z_{c}}^{(i)})^{2})\bigg] \\
&+ \frac{1}{L}\sum_{l=1}^{L} ||x^{(i,l)} - \hat{x}^{(i,l)}||_{2} + \frac{1}{L}\sum_{l=1}^{L} ||c^{(i,l)} - \hat{c}^{(i,l)}||_{2} + \lambda \sum_{w}|w| = \\
&- \frac{1}{2}\sum_{i=1}^{|z_{x}|}   (1 + log((\sigma_{z_{x}}^{(i)})^{2}) - (\mu_{z_{x}}^{(i)})^{2} - (\sigma_{z_{x}}^{(i)})^{2}) - \sum_{i=1}^{|z_{c}|}   (1 + log((\sigma_{z_{c}}^{(i)})^{2}) - (\mu_{z_{c}}^{(i)})^{2} - (\sigma_{z_{c}}^{(i)})^{2})] \\
&+ \frac{1}{L}\sum_{l=1}^{L} ||x^{(i,l)} - \hat{x}^{(i,l)}||_{2} + \frac{1}{L}\sum_{l=1}^{L} ||c^{(i,l)} - \hat{c}^{(i,l)}||_{2} + \lambda \sum_{w}|w|
\end{split}
\end{equation}

Where $L=1$ since a mini-batch size of 200 was used, $|z|$, $|z_{c}|$ and $|z_{x}|$ are the dimensions of the latent variables $z$, $z_{c}$ and $z_{x}$ respectively.
For optimization Adam \cite{journals/corr/KingmaB14} was employed. Note that the aforementioned architecture is likely not optimal and was chosen based on previous personal experience for illustrative purposes with no attempt to find an optimal hyper-parameter setting for this experiment. Training was performed with early stop strategy once the loss on the validation set has started increasing.

\subsubsection{Additional Baselines}
\label{S:Experimental Results:Kddcup99:Additional Baselines}
Despite aiming to compare primarily against the results presented in  \cite{DBLP:journals/corr/LiangP16} for diligence the same data set was evaluated by three additional algorithms: Isolation Forest \cite{Liu:2012:IAD:2133360.2133363}, One Class SVM \cite{Cortes1995} and Local Outlier Factor \cite{Breunig:2000:LID:335191.335388}. Not much effort was put into fine tuning these algorithms on the target data set and the results should be taken as indicative only.
\subsubsection{Metrics}
\label{S:Experimental Results:Kddcup99:Metrics}
The following metrics were evaluated against each of the methods:
\begin{enumerate}
\item Area under the Precision-Recall Curve (PRC): The area under the curve when plotting the recall on the x-axis against precision on the y-axis for all relevant possible threshold values for discriminating between normal and anomalous observations. PRC is recommended in scenarios where the data set is highly imbalanced \cite{Davis:2006:RPR:1143844.1143874}. The area under the curve (AUC) provides a summary statistic to the performance of a classifier in the PRC space.
\item Average Precision Score (APS): Provides a summary statistic for the Precision-Recall Curve as a weighted mean of precision obtained at each threshold, with the weight being the increase in recall from the previous threshold, calculated as $APS = \sum_{n}(R_{n}-R_{n-1})P_{n}$ where $P_{n}$ and $R_{n}$ are the precision and recall at the $n-th$ threshold.
\item Area under the Receiver Operating Characteristics Curve (ROC): The ROC curve enables the visualization of the relative trade-off between true-positive rate (TPR) and false-positive rate (FPR) by plotting the FPR on the x-axis against TPR on the y-axis for all relevant threshold values. The area under the curve (AUC) provides a summary statistic to the performance of a classifier in the ROC space.
\item Top-100 Precision: The fraction of correctly detected anomalies  in the top 100 scored observations.
\end{enumerate}

\subsubsection{Performance Metrics for Standard Classification}
\label{S:Experimental Results:Performance Metrics for Standard Classification}
Due to the challenges related to binary classification over a highly imbalanced data sets \cite{DBLP:journals/corr/BrancoTR15} a cross-validation with 5-fold stratified partitioning was performed where the ratio of the two classes in each of the the train/test partitions was kept equal to the distribution in the complete data set. The results are summarized in the following tables:

\begin{table}[H]
\centering
\begin{tabular}{||c c c c c||} 
\hline
Method & PRC (AUC) & APS & ROC (AUC) & Top-100 Precis. \\ [0.5ex] 
\hline\hline
\textbf{JLVAE} & \textbf{0.51848} & \textbf{0.51874} & \textbf{0.99257} & 0.018 \\ 
\hline
IF \cite{Liu:2012:IAD:2133360.2133363} & 0.00842 & 0.00855 & 0.01937 & 0 \\
\hline
OCSVM \cite{Cortes1995} & 0.00846 & 0.00853 & 0.02459 & 0 \\
\hline
LOF \cite{Breunig:2000:LID:335191.335388} & 0.02458 & 0.03579 & 0.64849 & \textbf{0.056} \\
\hline
\end{tabular}
\caption{\label{tab:all means results}Summary of mean results obtained over the 5-folds for all methods.}
\end{table}

The results obtained demonstrate a substantial improvement compared to the benchmark algorithms tested in the described setting and to the results obtained by \cite{DBLP:journals/corr/LiangP16} for a similar data set. The following tables contain detailed information as to the results obtained for each of the algorithms and k-folds.

\begin{table}[H]
\centering
\begin{tabular}{||c c c c c||} 
\hline
K-Fold & PRC (AUC) & APS & ROC (AUC) & Top-100 Precision \\ [0.5ex] 
\hline\hline
1 & 0.50543 & 0.5057 & 0.99240 & 0.01 \\ 
\hline
2 & 0.53492 & 0.53524 & 0.99321 & 0.03 \\
\hline
3 & 0.51293 & 0.5131 & 0.99227 & 0.01 \\
\hline
4 & 0.53134 & 0.53162 & 0.99264 & 0.03 \\
\hline
5 & 0.50777 & 0.50805 & 0.99233 & 0.01 \\
\hline\hline
 mean & 0.51848 & 0.51874 & 0.99257 & 0.018 \\
\hline
\end{tabular}
\caption{\label{tab:jlvae results}JLVAE - proposed method.}
\end{table}

\begin{table}[H]
\centering
\begin{tabular}{||c c c c c||} 
\hline
K-Fold & PRC (AUC) & APS & ROC (AUC) & Top-100 Precision \\ [0.5ex] 
\hline\hline
1 & 0.0084 & 0.00854 & 0.01773 & 0 \\ 
\hline
2 & 0.0084 & 0.0085 & 0.01203 & 0 \\
\hline
3 & 0.00843 & 0.00859 & 0.02282 & 0 \\
\hline
4 & 0.00848 & 0.0086 & 0.02942 & 0 \\
\hline
5 & 0.00838 & 0.00853 & 0.01486 & 0 \\
\hline\hline
 mean & 0.00842 & 0.00855 & 0.01937 & 0 \\
\hline
\end{tabular}
\caption{\label{tab:if results}Isolation Forest.}
\end{table}

\begin{table}[H]
\centering
\begin{tabular}{||c c c c c||} 
\hline
K-Fold & PRC (AUC) & APS & ROC (AUC) & Top-100 Precision \\ [0.5ex] 
\hline\hline
1 & 0.00847 & 0.00854 & 0.02476 & 0 \\ 
\hline
2 & 0.00847 & 0.00853 & 0.02475 & 0 \\
\hline
3 & 0.00846 & 0.00853 & 0.02462 & 0 \\
\hline
4 & 0.00846 & 0.00853 & 0.02469 & 0 \\
\hline
5 & 0.00846 & 0.00852 & 0.02414 & 0 \\
\hline\hline
 mean & 0.00846 & 0.00853 & 0.02459 & 0 \\
\hline
\end{tabular}
\caption{\label{tab:ocsvm results}One Class SVM.}
\end{table}

\begin{table}[H]
\centering
\begin{tabular}{||c c c c c||} 
\hline
K-Fold & PRC (AUC) & APS & ROC (AUC) & Top-100 Precision \\ [0.5ex] 
\hline\hline
1 & 0.02442 & 0.03593 & 0.64976 & 0.01 \\ 
\hline
2 & 0.02464 & 0.03627 & 0.64744 & 0.09 \\
\hline
3 & 0.02394 & 0.0353 & 0.64671 & 0.03 \\
\hline
4 & 0.02483 & 0.0351 & 0.64564 & 0.08 \\
\hline
5 & 0.02506 & 0.03633 & 0.65290 & 0.07 \\
\hline\hline
 mean & 0.02458 & 0.03579 & 0.64849 & 0.056 \\
\hline
\end{tabular}
\caption{\label{tab:lof results}Local Outlier Factor.}
\end{table}

\subsection{Waste Water Treatment Plant}
\label{S:Waste Water Treatment Plant}

\subsubsection{Robustness to Contextual Anomalies}
\label{S:Experimental Results:Robustness to Contextual Anomalies}
To demonstrate the effectiveness of the method in dealing with contextual anomalies it was evaluated on a real-world waste water treatment plant located at Western Australia. The plant design features a splitter chamber that divides the incoming waste water into two wells each having two pumps. Waste water pumped by the pumps are then merged into a single outlet pipe by a series of two joiner pipes, one joining the pumps output in each well, and one joining the two well's output. The control logic for the plant under normal conditions will turn pumps on and off as required to meet inflow conditions and also use variable speed drives to modulate the speed of the operational pumps based on a level reading of the splitter chamber. This design results in a system where the operational characteristics of a pump is not independent from the other pumps. 
\subsubsection{Data}
\label{S:Experimental Results:Waste Water Treatment Plant:Data}
The data set contains roughly 30 months of operational data, close to 150 attributes and about 690,200 coincident observations with 2 minutes frequency and is comprised of the following information:

\begin{enumerate}
\item Sensors specific per pump such as vibration, temperature, speed, operational pressures and flows, power supply characteristics, and more.
\item Generated features per pump such as efficiency.
\item Environmental readings from the two wells.
\item Other useful data such as the splitter chamber level and external weather conditions. 
\end{enumerate}

The data is assumed to contain anomalies of unknown nature and frequency. The data was partitioned such that data generated in a particular pump run-cycle was kept together and not partitioned across sets. Partitioning was done into training (65\%), validation(15\%) and testing (\%20) sets where the percentages represent the portion of pump run-cycles rather than single observations. Lastly the data was not pre-processed except for aligning observations in time by mean interpolation, discarding partial observations with the remaining observations standardized. Note that there are no categorical attributes in this data set.

\subsubsection{Model}
\label{S:Experimental Results:Waste Water Treatment Plant:Model}
A model is developed for each pump individually where the behavioral attribute are the data relating directly to the operational sensor readings of the pump, and where contextual attributes are some of the behavioral attributes of the remaining pumps as well as environmental factors such as the splitter chamber level and weather conditions. For example, a model for pump one will include as context the inflow and outflow rate and pressure of pumps 2-4, the splitter chamber level and environmental information. The setup was similar to the one described in section \ref{S:Experimental Results:Kddcup99:Model} with arrangement of units in the behavioral recognizer as follows: 28 (input), 20, 10 and 5 units for the latent output, with the generator having a mirror architecture. The arrangement of units in the contextual recognizer were: 38 (input), 20, 10 and 2 units for the latent output, with the generator having 4, 7 and 10 units in the output layer. Note that similarly to the previous experiment the aforementioned architecture is likely not optimal and was chosen based on personal experience of the author for illustrative purposes.

\subsubsection{Metrics}
\label{S:Experimental Results:Waste Water Treatment Plant:Metrics}
In this case it is intended to evaluate the models robustness to contextual anomalies and novelties. To do so the following method is applied. A threshold was set so that the number of anomalies detected by the model in the test data set is roughly 1\%. Then 10,000 normal observation are randomly selected and transformed by scaling and offsetting a randomly chosen subset of the attributes element-wise where $scale \sim \mathcal{U}(-2.5, 2.5)$ and $offset \sim \mathcal{U}(-2.0, 2.0)$ resulting in 15 new test data sets. For the A data sets approximately 10\% of each group was transformed, for the B data sets about 30\% and for the C data sets about 50\%. For the D,E and F data sets the same absolute number of attributes was transformed in each of the groups. Given the attributes are standardized to zero mean and unit standard deviation the noise levels applied to the attributes are substantial and result in many anomalies detected as per the summary in table \ref{tab:noisy anomalies}:

\begin{table}[H]
\centering
\begin{tabular}{||c c c c||}
\hline
Data set & \# behavior trans. & \# context trans. & \# anomalies reported \\ [0.5ex] 
\hline\hline
A1 & 3 & 0 & 1240 \\ 
\hline
A2 & 0 & 1 & 7 \\ 
\hline
A3 & 3 & 1 & 576 \\ 
\hline
B1 & 9 & 0 & 4614 \\ 
\hline
B2 & 0 & 3 & 2 \\ 
\hline
B3 & 9 & 3 & 4251 \\ 
\hline
C1 & 14 & 0 & 6949 \\ 
\hline
C2 & 0 & 5 & 10 \\ 
\hline
C3 & 14 & 5 & 5909 \\ 
\hline
Dx & 2 & 0 & 288 \\ 
\hline
Dc & 0 & 2 & 0 \\ 
\hline
Ex & 5 & 0 & 1567 \\ 
\hline
Ec & 0 & 5 & 13 \\ 
\hline
Fx & 10 & 0 & 4485 \\ 
\hline
Fc & 0 & 10 & 17 \\ 
\hline

\hline
\end{tabular}
\caption{\label{tab:noisy anomalies}Summary of number of anomalies detected in the noisy data sets.}
\end{table}

The results demonstrate the algorithm is robust to anomalies and novelties in the contextual data attributes whilst maintaining sensitivity to anomalies in the behavioral space. It is notable that even when the entire set of contextual attributes is transformed in data set Fc, still less anomalies are reported than data set Dx where only two behavioral attributes are corrupted with noise.

\section{Conclusion}
In this paper a novel algorithm for contextual anomaly detection is presented and a novel ANN architecture comprised of multiple cross-linked VAEs to model directed graphical distribution models for modeling generative processes. The algorithm performs well in the test scenarios and is robust to contextual anomalies and novelties.

\section{Acknowledgements}
This research was supported by the Water Corporation of Western Australia. I gratefully acknowledge my colleagues from the Water Corporation for access to infrastructure and for their cooperation, which greatly assisted the research.

\bibliographystyle{abbrv}
\bibliography{refs}

\end{document}